\documentclass{article}

\usepackage{graphicx}%
\usepackage{multirow}%
\usepackage{amsmath,amssymb,amsfonts}%
\usepackage{amsthm}%
\usepackage{mathrsfs}%
\usepackage[title]{appendix}%
\usepackage{textcomp}%
\usepackage{manyfoot}%
\usepackage{booktabs}%
\usepackage{algorithm}%
\usepackage{algorithmicx}%
\usepackage{algpseudocode}%
\usepackage{listings}%
\usepackage{caption}
\usepackage{multicol}
\usepackage{bm}
\usepackage{todonotes}
\usepackage{changepage}
\usepackage{listings}
\usepackage{hhline}
\usepackage{url}

\graphicspath{ {images/} }

\begin{document}

\title{Optimal Weighted Convolution for Classification and Denosing}

\makeatletter
\onecolumn
{\fontsize{18pt}{20pt}\selectfont\bfseries\@title\par}
Simone Cammarasana
 \footnote{
\textbf{Simone Cammarasana}
CNR-IMATI, Via De Marini 6, Genova, Italy \\
simone.cammarasana@ge.imati.cnr.it
}
  , 
 Giuseppe Patan\`e
  \footnote{
 \textbf{Giuseppe Patan\`e} 
 CNR-IMATI, Via De Marini 6, Genova, Italy 
}

\makeatother

\abstract{We introduce a novel weighted convolution operator that enhances traditional convolutional neural networks (CNNs) by integrating a spatial density function into the convolution operator. This extension enables the network to differentially weight neighbouring pixels based on their relative position to the reference pixel, improving spatial characterisation and feature extraction. The proposed operator maintains the same number of trainable parameters and is fully compatible with existing CNN architectures. Although developed for 2D image data, the framework is generalisable to signals on regular grids of arbitrary dimensions, such as 3D volumetric data or 1D time series. We propose an efficient implementation of the weighted convolution by pre-computing the density function and achieving execution times comparable to standard convolution layers. We evaluate our method on two deep learning tasks: image classification using the CIFAR-100 dataset~\cite{krizhevsky2009learning} and image denoising using the DIV2K dataset~\cite{Agustsson2017CVPRWorkshops}. Experimental results with state-of-the-art classification (e.g., VGG~\cite{simonyan2015a}, ResNet~\cite{he2016deep}) and denoising (e.g., DnCNN~\cite{zhang2017beyond}, NAFNet~\cite{chen2022simple}) methods show that the weighted convolution improves performance with respect to standard convolution across different quantitative metrics. For example, VGG achieves an accuracy of~$66.94\%$ with weighted convolution versus~$56.89\%$ with standard convolution on the classification problem, while DnCNN improves the PSNR value from 20.17 to 22.63 on the denoising problem. All models were trained on the CINECA Leonardo cluster to reduce the execution time and improve the tuning of the density function values. The PyTorch implementation of the weighted convolution is publicly available at: \url{https://github.com/cammarasana123/weightedConvolution2.0}.}

\section{Introduction}\label{sec:INTRO}
Deep learning (DL) is a subset of artificial intelligence methodologies that focuses on the processing and analysis of signals defined on regular grids (e.g., 2D images). DL accounts for large datasets and hierarchical feature representations, including multiple linear and non-linear layers. These transformations map the input signal into higher-level abstractions, reducing the number of parameters required to represent the signal. DL is widespread in image-based applications, such as robotics~\cite{soori2023artificial}, autonomous navigation~\cite{krishnan2021air}, computational chemistry~\cite{goh2017deep}, and healthcare~\cite{cammarasana2022real}. The convolution operator is applied to 2D images to extract both geometrical and texture features and characterise spatial hierarchies and local patterns, leading to a class of DL models named convolutional neural networks (CNNs). In standard CNNs, convolution operations equally scale neighbouring pixels of a reference pixel, relying on optimised kernel weights to determine feature relevance. This assumption implies that all pixels in a local region contribute equally to the convolution, independent of their relative position (Sect.~\ref{sec:BGRW}).

We propose a novel weighted convolution operator that enhances the standard convolutional framework by incorporating a spatial density function, allowing the network to account for the relative position of each pixel and improving spatial characterisation and feature extraction. Although our method is focused on 2D image data, the weighted convolution framework is general to any signal defined on regular grids, such as 3D volumetric data (e.g., medical imaging), 1D time-series signals, or spatio-temporal sequences. As a significant advantage, the proposed weighted convolution does not increase the number of trainable parameters with respect to standard convolution, and is fully compatibility with existing convolutional neural network frameworks (see Sect.~\ref{sec:METHOD}). We propose an efficient implementation of the weighted convolution with a pre-computation of the density function and the update of the kernel weights with the density function at every iteration, reducing the computational cost with an execution time comparable to the standard convolution layers.

We analyse the application of our weighted convolution to state-of-the-art deep learning problems and architectures. In particular, we discuss two different problems: (i) a classification problem with the CIFAR-100, a dataset of 60K images with~$32 \times 32$ resolution and 100 classes; (ii) a denoising problem with the DIV2K, a dataset of 1,000 high-resolution RGB images. For the problem (i), we compare five different state-of-the art classification methods: the \emph{deep residual learning for image recognition} (ResNet), \emph{very deep convolutional networks for large-scale image recognition} (VGG), \emph{network in network} (NIN), \emph{gated multi layer perceptron} (gMLP), and \emph{gated attention coding for spiking neural networks} (GAC-SNN), discussing the classification results through quantitative metrics: accuracy, F1-score, and confusion matrix. A sixth method, i.e., EfficientNet, has been tested without achieving convergence with both standard and weighted convolution. For the problem (ii), we compare three different state-of-the-art denoising methods: the \emph{residual learning of deep CNN} (DnCNN), \emph{nonlinear activation free network} (NAFNET), and \emph{cascaded gaze network} (CGNet), discussing the denoising results through quantitative metrics: \emph{peak signal-to-noise ratio} (PSNR), \emph{structural similarity} (SSIM), \emph{feature-based similarity} (FSIM), \emph{normalised mean squared error} (MSE), and \emph{universal image quality} (UIQ). For both problems (i) and (ii), we compare the standard convolution with our weighted convolution within the learning architectures. In particular, we manage the density function values as hyperparameters of the networks, keeping the same number of training parameters (Sect.~\ref{sec:EXPRES}).

Our weighted convolution improves the results of standard convolution for all the tested methods. In the classification problem, the VGG has an accuracy of~$66.94\%$ with the weighted convolution, compared to an accuracy of~$56.89\%$ with the standard convolution. In the denoising problem, DnCNN has a PSNR value of~$22.63$ with the weighted convolution, compared to a PSNR value of~$20.17$ with the standard convolution. For the denoising problem, we also compare the results with different kernel sizes, showing that a~$5 \times 5$ kernel and a density function with a small contribution of the external elements have better results than a~$3 \times 3$ kernel. For the classification problem, we apply only~$3 \times 3$ kernel size, due to the small image resolution. We have performed our tests on the CINECA Leonardo cluster. The parallelisation and high-performance of the hardware allow us to tune the density function to improve the accuracy of the weighted convolution and the results of the classification and denoising.

We underline that we have implemented the public versions of all the learning methods for both classification and denoising. For a fair comparison, we have defined the same hyper-parametrisation in terms of weights initialisation, optimisation criteria, and data augmentation. For this reason, the accuracy results slightly differ with respect to the results mentioned in the respective papers. Finally, we discuss conclusions and future work (Sect.~\ref{sec:CONC}) as the extension to 3D kernels and applicative contexts (e.g., biomedical data).
The PyTorch class for the weighted convolution is available at \url{https://github.com/cammarasana123/weightedConvolution2.0}.

\section{Related work}\label{sec:BGRW}

\paragraph*{Convolutional neural network}
The \emph{weighted convolutional neural network ensemble}~\cite{frazao2014weighted} combines output probabilities from multiple CNNs, assigning greater influence to models with higher accuracy. The \emph{constrained convolution layer}~\cite{abbood2022new} restricts the number of trainable weights within each kernel by pruning less significant connections during training. \emph{DropOut}~\cite{wan2013regularization} reduces overfitting through a regularisation technique where each unit in a fully connected layer is randomly deactivated with a probability~$(1-p)$. Additional regularisation strategies~\cite{santos2022avoiding, hou2019weighted, zheng2018improvement} further improve generalisation. Hyperparameter optimisation is relevant for sensor-based human activity recognition~\cite{raziani2022deep}, prediction of Parkinson’s disease~\cite{kaur2020hyper}, and emotion recognition in intelligent tutoring systems~\cite{zatarain2020hyperparameter}. Self-attention mechanisms~\cite{vaswani2017attention} dynamically focus on different parts of the input, improving the characterisation of global dependencies.

In~\cite{jia2022variable}, convolutional kernels incorporate an additional weighted parameter for each kernel element, thus increasing the trainable variables of the model. In \emph{dynamic convolution}~\cite{chen2020dynamic}, multiple kernels are dynamically weighted according to the input and adapt the convolution operation. \emph{Omni-dimensional dynamic convolution}~\cite{li2022omni} accounts for a multi-dimensional attention mechanism to learn complementary attentions for convolutional kernels along the dimensions of the kernel tensor \emph{Generalised convolution}~\cite{ghiasi2019generalizing} replaces the traditional inner product with positive definite kernel functions (e.g., Gaussian, Laplacian). In \emph{convolutional kernel networks}~\cite{mairal2014convolutional}, local kernel approximations are applied to spatial regions of the input data to learn adaptive kernel functions. This approach has been extended to graph-based data~\cite{chen2020convolutional}, where kernels represent graph structure through local sub-patterns.

\paragraph*{Deep learning methods for classification}
The \emph{very deep convolutional networks for large-scale image recognition} (VGG)~\cite{simonyan2015a} architecture introduces a deep convolutional neural network with small convolutional filters, increasing depth with simple and uniform layers. The \emph{deep residual learning for image recognition} (ResNet)~\cite{he2016deep} applies residual learning through skip connections to reduce the vanishing gradient problem. The \emph{Mobilenetv2}~\cite{sandler2018mobilenetv2} introduces an inverted residual structure where the shortcut connections are among the bottleneck layers and the intermediate expansion layer applies lightweight depthwise convolutions to filter features as a source of non-linearity \emph{Network in network} (NiN)~\cite{lin2013network} replaces traditional convolutional layers with small neural networks to capture complex features within local receptive fields. The \emph{gated multi-layer perceptron} (gMLP)~\cite{liu2021pay} applies gated MLP layers with spatial modulating units to characterise dependencies without attention mechanisms. The \emph{gated attention coding for spiking neural networks} (GAC-SNN) introduces a novel group attention clustering to represent spatial correlations and shared features across neuron groups, accounting for unsupervised clustering to increase sparsity and robustness~\cite{qiu2024gated}. \emph{EfficientNet}~\cite{tan2019efficientnet} uniformly scales network dimensions using a fixed set of scaling coefficients.

\paragraph*{Deep learning methods for denoising}
The \emph{residual learning of deep CNN} (DnCNN)~\cite{zhang2017beyond} introduces a residual learning strategy combined with batch normalisation to remove Gaussian noise across various noise levels. The \emph{fast and flexible denoising convolutional neural network} (FFDNet)~\cite{zhang2018ffdnet} addresses a noise level map as input, accounting for spatially variant noise and improving flexibility during inference. The \emph{restoration transformer}~\cite{zamir2022restormer} proposes a multi-stage transformer with a gated self-attention mechanism adapted for low-level vision through a hierarchical encoder-decoder structure and feature aggregation. The \emph{nonlinear activation free network} NAFNet~\cite{chen2022simple} introduces a lightweight attention-free network architecture using simple gated convolutions and normalisation-free layers. The \emph{cascaded gaze network} (CGNet)~\cite{ghasemabadi2024cascadedgaze} is an encoder-decoder architecture that removes self-attention mechanisms, applies a cascading structure, and progressively aggregates local features to capture global context. Singular value decomposition is applied through the learning of the optimal threshold values~\cite{cammarasana2025analysis}. In the \emph{Noise2Noise} algorithm~\cite{lehtinen2018noise2noise}, the network learns to denoise images accounting for the noisy data without any knowledge of the ground-truth. The \emph{Noise2Void} algorithm~\cite{krull2019noise2void} further expands this idea without requiring a couple of noisy images for the training. This approach is relevant in biomedical fields, where ground-truth images are typically not available. The \emph{Noise2Self} method~\cite{batson2019noise2self} proposes a self-supervised algorithm without requiring any prior assumption on the input image and estimation of the noise.

\section{Our weighted convolution~$\&$ parameters' tuning}\label{sec:METHOD}
After recalling the standard convolution (Sect.~\ref{sec:CONV}), we introduce the weighted convolution and the efficient computation of the density function values (Sect.~\ref{sec:PARAM}).

\subsection{Standard convolution operator}\label{sec:CONV}
Given a compact domain~$\Omega \in \mathbb{R}^n$ and two functions~$f, g \in \mathcal{L}^1(\Omega)$, the convolution is defined as
\begin{equation}\label{eq:convCont}
 (f \ast g) (\bm{t}) := \int_{\Omega} f(\bm{\tau}) g(\bm{t}-\bm{\tau}) d\bm{\tau}.
 \end{equation}
In CNNs,~$\Omega$ is a discrete 2D domain and the functions~$f,g$ are the input signal (e.g., the 2D image) and the filter (e.g., the convolution kernel), respectively. Given the input signal~$\bm{I} \in \mathbb{R}^{R \times C}$ defined on a 2D regular grid and the tensor of kernels~$\bm{W} \in \mathbb{R}^{K_a \times K_b \times F}$ composed of F kernels~$\bm{w}$ of size~$K_a \times K_b$ as~$(\bm{W})_{ijf} := \bm{w}_{ij}^f~$, we discretise the convolution in Eq.~(\ref{eq:convCont}) of~$\bm{I}$ with the kernels~$\bm{W}$ as
\begin{equation}\label{eq:CONVOLUTION}
\left\{
\begin{array}{l}
(\bm{I} \ast \bm{W})_{ij}^f := \sum_{a=1}^{K_a} \sum_{b=1}^{K_b} \bm{w}_{ab}^{f} \cdot \bm{I}_{i+a-K_a+1,j+b-K_b+1},\\
i = 1 \ldots R, j = 1\ldots C, f = 1\ldots F.
\end{array}
\right.
\end{equation}
Introducing the neighbourhood~$\mathcal{N}(\bm{I}_{ij})$ as the~$K_a \times K_b$ sub-matrix of~$\bm{I}$ centred in~$(i,j)$, the discrete convolution is rewritten in matrix form as \mbox{$(\bm{I} \ast \bm{W})_{ij}^f := \langle \bm{w}^f , \mathcal{N}(\bm{I}_{ij}) \rangle_F$}, where
the Frobenius inner product~$\langle \cdot, \cdot \rangle_F$ of two matrices~$\mathbf{A}, \mathbf{B} \in \mathbb{R}^{m \times n}$ is defined as~$\langle \mathbf{A}, \mathbf{B} \rangle_F := \sum_{i=1}^{m} \sum_{j=1}^{n} A(i,j) B(i,j)$.

Given an input~$\mathcal{I} = \left\{\bm{I}_i\right\}_{i=1}^N$ and target~$\mathcal{T} = \left\{\bm{T}_i\right\}_{i=1}^N$ data set of~$N$ images, we define a learning model~$\mathcal{M}$ as the minimisation of a loss function~$\mathcal{L}$ with respect to the kernels~$\bm{W}$ (i.e., the weights) between the output data set and the output of the network~$\hat{\mathcal{T}}$
\begin{equation}\label{eq:learningModel}
\mathcal{M}:\qquad \min_{\bm{W}} \mathcal{L}( \mathcal{T} , \hat{\mathcal{T}}(\bm{W}) ),
\end{equation}
where~$\hat{\mathcal{T}}$ is the output of the combination of convolution layers in Eq.~(\ref{eq:CONVOLUTION}) and non-linear operators applied to the input~$\mathcal{I}$.

\subsection{Weighted convolution operator\label{sec:WCONV}}
Given the density function~$\varphi \in \mathcal{L}^{\infty}(\Omega)$, we introduce the \emph{weighted convolution} as \mbox{$(f \ast g_\varphi)(\bm{t}) := \int_{\Omega}  f(\bm{\tau}) (\varphi(\bm{t}-\bm{\tau})  g(\bm{t}-\bm{\tau})) d\bm{\tau}$}. The weighted convolution reduces to the standard convolution when \mbox{$\varphi:=1$}. Discretising the density function~$\varphi$ on a 2D regular grid as~$\bm{\Phi} \in \mathbb{R}^{K_a \times K_b}$, we define the tensor of kernels with density function as \mbox{$\bm{W}_{\bm{\Phi}} \in \mathbb{R}^{K_a \times K_b \times F}$}, where~$(\bm{W}_{\bm{\Phi}})_{ijf} := \bm{\Phi}_{ij} \bm{w}_{ij}^f~$, and the \emph{discrete weighted convolution} as
\begin{equation}\label{eq:wCONVOLUTION}
\left\{
\begin{array}{l}
(\bm{I} \ast \bm{W}_{\bm{\Phi}})_{ij}^{f} := \sum\limits_{a=1}^{K_a} \sum\limits_{b=1}^{K_b} (\bm{\Phi}_{ab} \bm{w}_{ab}^{f} ) \bm{I}_{i+a-K_a+1,j+b-K_b+1},\\
i = 1 \ldots R,\, j = 1\ldots C,\, f = 1\ldots F.
\end{array}
\right.
\end{equation}
In matrix form, \mbox{$(\bm{I} \ast \bm{W}_{\bm{\Phi}})_{ij}^{f} := \langle \bm{\Phi} \circ \bm{w}^f, \mathcal{N}(\bm{I}_{ij} )\rangle_F$}, where the Hadamard product~$\circ$ is the component-wise product of two matrices, i.e.,~$\langle\mathbf{A},\mathbf{B}\rangle_{F}=:\mathbf{C}$,~$C(i,j):=A(i,j)B(i,j)$,~$\mathbf{A},\,\mathbf{B},\,\mathbf{C}\in\mathbb{R}^{m\times n}$. The density function~$\bm{\Phi}$ is shared across both the image and the kernels. We underline that the discrete weighted convolution with a uniform density function~$\bm{\Phi} = \bm{1}$ is equivalent to the discrete convolution without a density function. Table~\ref{tab:convComparison} compares standard and weighted convolution.
\begin{table}[t]
\caption{Comparison between standard and weighted convolution.\label{tab:convComparison}}
\begin{tabular}{c}
\includegraphics[width=1.0\textwidth]{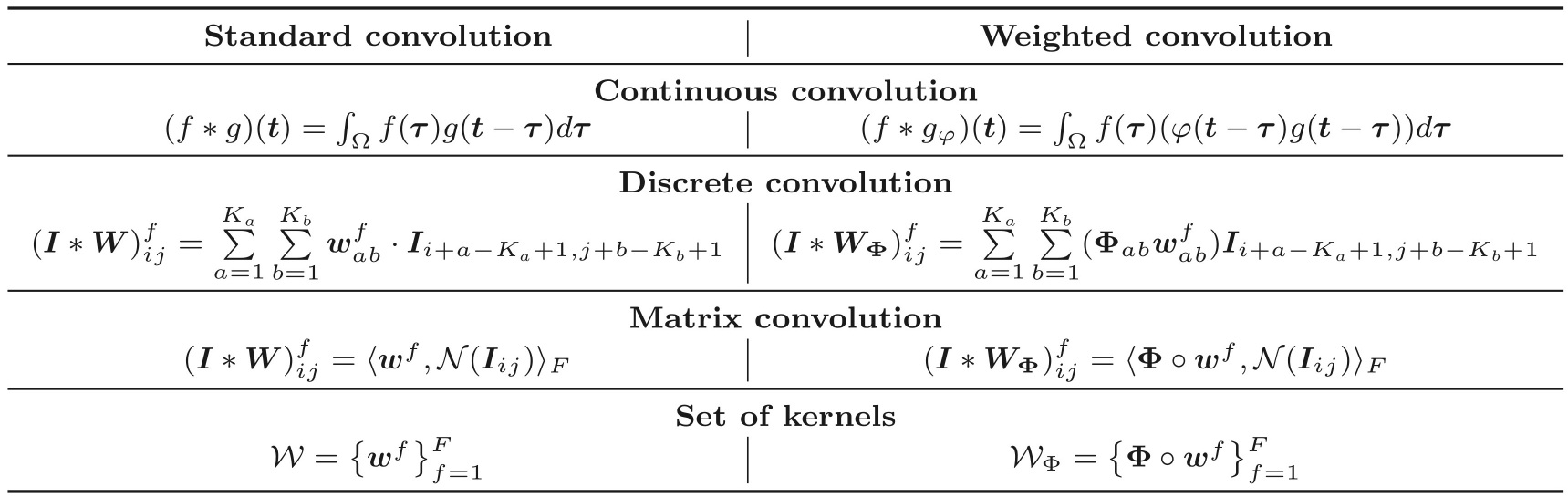}
\end{tabular}
\end{table}
Given the learning model in Eq.~(\ref{eq:learningModel}) and replacing the standard convolution~$\bm{I} \ast \bm{W}$ in Eq.~(\ref{eq:CONVOLUTION}) with the convolution with the density function~$\bm{I} \ast \bm{W}_{\bm{\Phi}}$ in Eq.~(\ref{eq:wCONVOLUTION}), we define the learning model
\begin{equation}\label{eq:wModel}
\mathcal{M}_{\bm{\Phi}}:\qquad\min_{\bm{W}} \mathcal{L}( \mathcal{T} , \hat{\mathcal{T}}(\bm{W}_{\bm{\Phi}}) ).
\end{equation}
\paragraph*{Efficient computation of the density function}\label{sec:PARAM}
Given a squared kernel~$K_a = K_b = K$,~$K$ odd,~$m = (K + 1)/2$, we define the discrete density function as the~$K_a \times K_b$ matrix~$\bm{\Phi} = \bm{\alpha} \bm{\beta}^{\top}$ (i.e.,~$ \bm{\Phi}_{ij} = \bm{\alpha}_i \bm{\beta}_j$), with~$\bm{\alpha} \in \mathbb{R}^{K_a}, \bm{\beta} \in \mathbb{R}^{K_b}$, where~$\bm{\alpha}$ and~$\bm{\beta}$ represent the scaling factors along the two dimensions of the regular grid. We assume the following properties for the density function:
\begin{itemize}
\item Symmetry along both the dimensions, i.e.,~$\bm{\alpha} = \bm{\beta}$,~$\bm{\alpha}(i) = \bm{\alpha}(K - i + 1), i = 1\ldots K$;
\item Value of the central node as~$\bm{\alpha}_m = M$.
\end{itemize}
The density function~$\bm{\Phi} \in \mathbb{R}^{K \times K}$ is defined through~$(K-1)/2$ coefficients, is symmetric, positive semidefinite, and with rank~$1$. Henceforth, we reduce the computation of the density function to the coefficients of~$\bm{\alpha}$, as~$\bm{\Phi} = \bm{\alpha} \bm{\alpha}^{\top}$. With our definition of the density function properties, a~$3 \times 3$ kernel size requires only one parameter for the density function, as~$\bm{\alpha} \in \mathbb{R}^3$,~$\bm{\beta} = \bm{\alpha}$,~$\bm{\alpha}_2 = M$, and~$\bm{\alpha}_1 = \bm{\alpha}_3$. Generally, it requires~$(K-1)/2$ parameters.
\begin{table}[t]
\caption{Weighted convolution class\label{sec:APPENDIX}}
\begin{tabular}{c}
\includegraphics[width=0.8\textwidth]{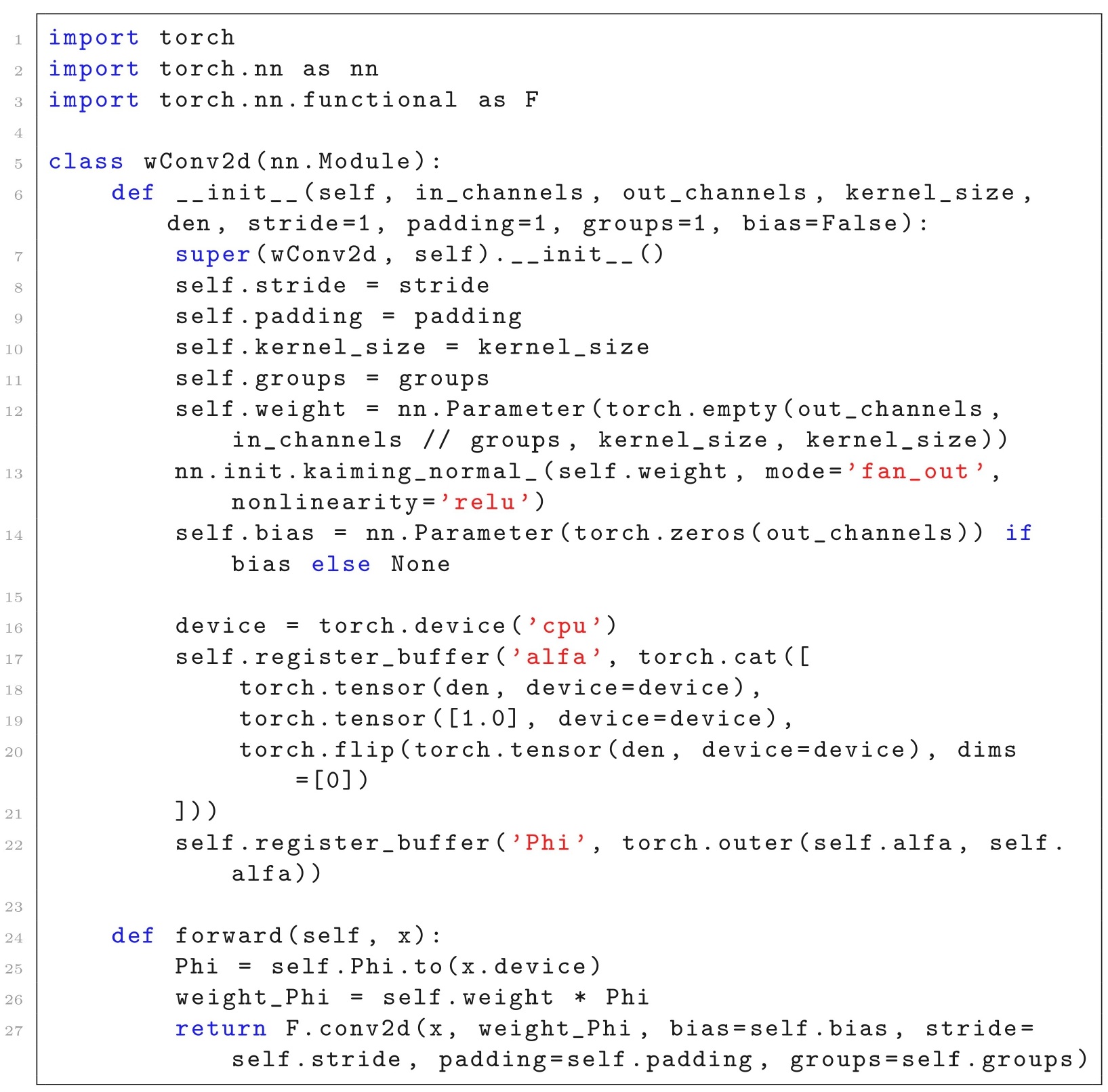}
\end{tabular}
\end{table}
%

The efficient implementation of the weighted convolution (Table~\ref{sec:APPENDIX}) pre-computes the density function~$\bm{\Phi}$ as an initialisation step, and updates the tensor of kernels with density function~$\bm{W}_{\bm{\Phi}}$ at every iteration. The proposed implementation maintains the same number of trainable parameters of standard convolution and is fully compatible with existing CNN architectures. Given an input image of size~$C \times N \times N$ (with~$C$ channels and~$N \times N$ resolution) and~$F$ filters of size~$K \times K$, the computational cost of the standard convolution is~$\mathcal{O}(N^2 \times C \times F \times K^2)$. The computational cost of the weighted convolution is~$\mathcal{O}(N^2 \times C \times F \times K^2 + K^2 \times F)$, where the computation of the kernels with density function adds~$K^2 \times F$ operations.
\begin{figure*}[t]
\centering
\begin{tabular}{cc|cc}
\multicolumn{4}{c}{Confusion matrix: Standard (left) and weighed (right) convolution}\\
\hline
\multicolumn{2}{c|}{ResNet 56} &\multicolumn{2}{c}{VGG}\\
\includegraphics[width=.22\textwidth]{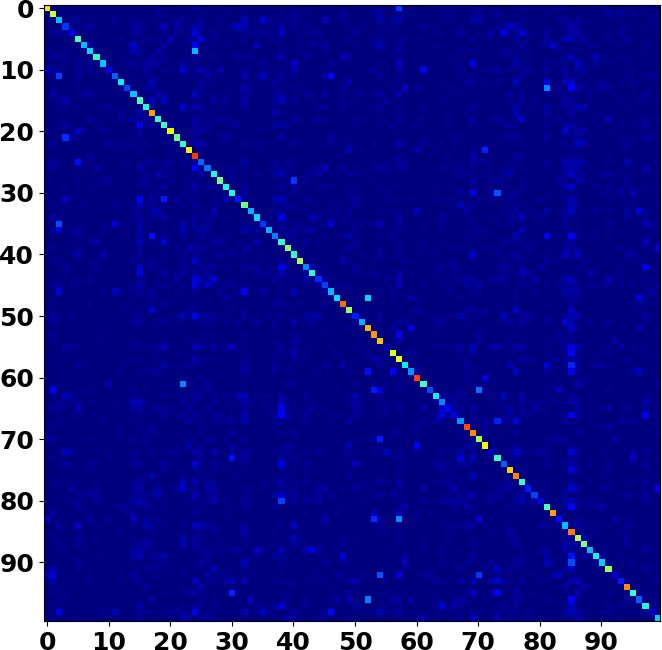}
&\includegraphics[width=.22\textwidth]{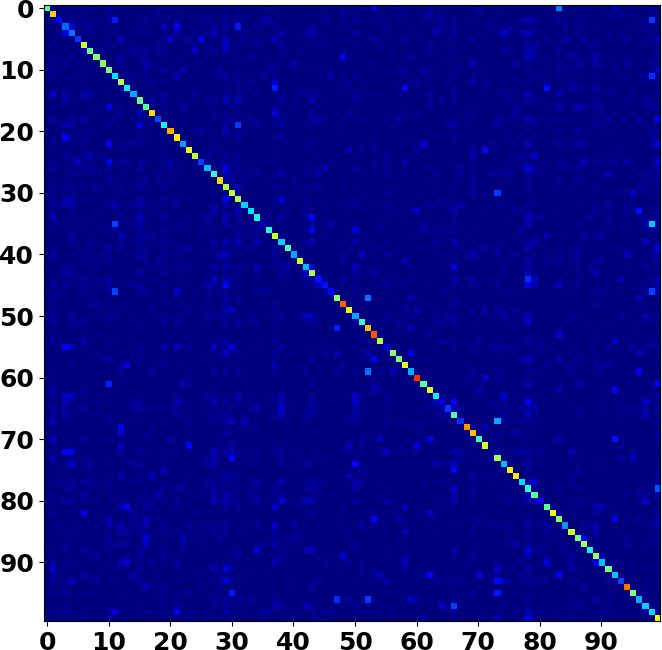} 
&\includegraphics[width=.22\textwidth]{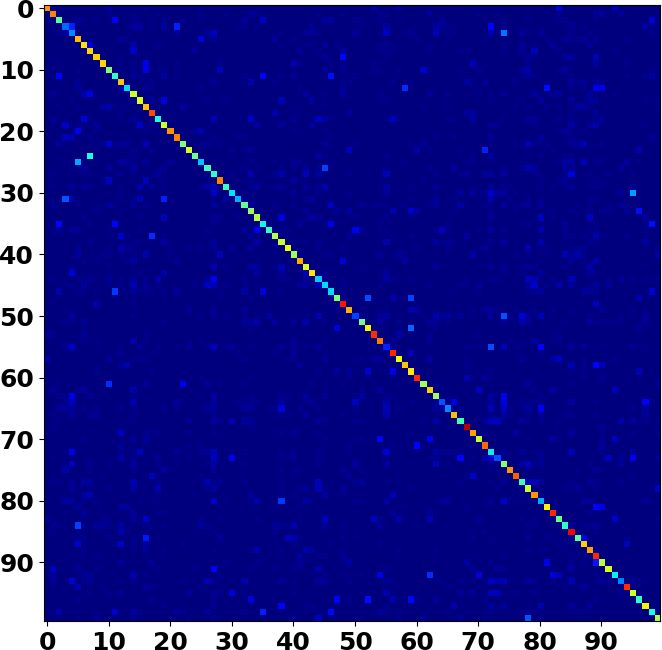}
&\includegraphics[width=.22\textwidth]{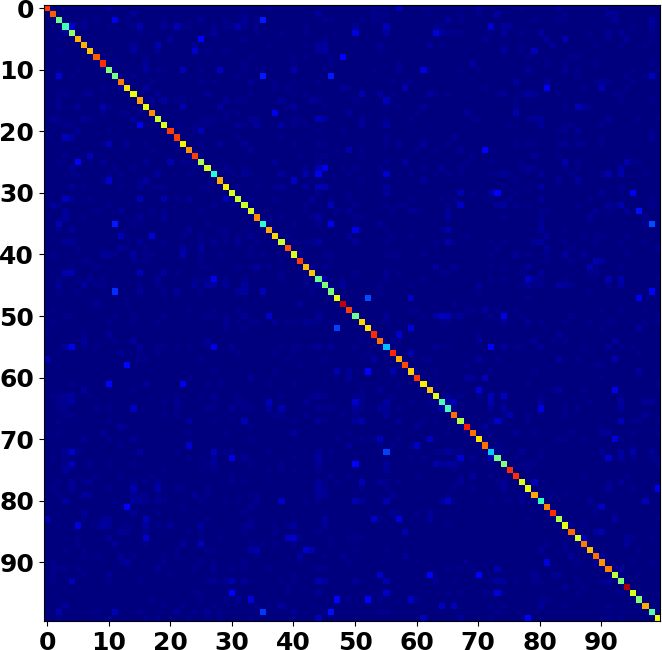} \\
\hline
\multicolumn{2}{c|}{NiN} &\multicolumn{2}{c}{gMLP}\\
\includegraphics[width=.22\textwidth]{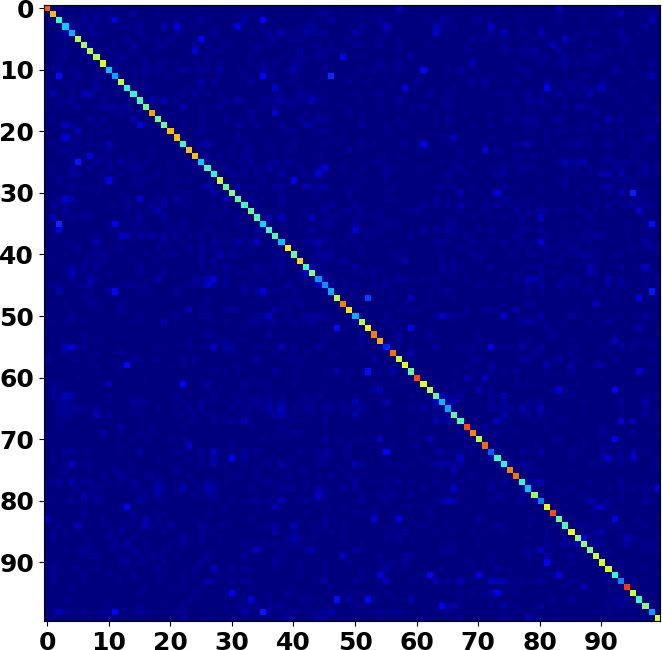}
&\includegraphics[width=.22\textwidth]{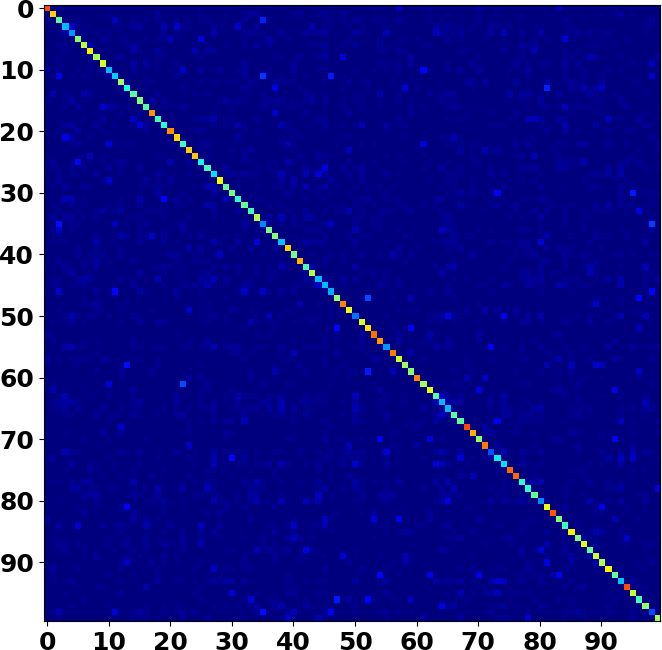}
&\includegraphics[width=.22\textwidth]{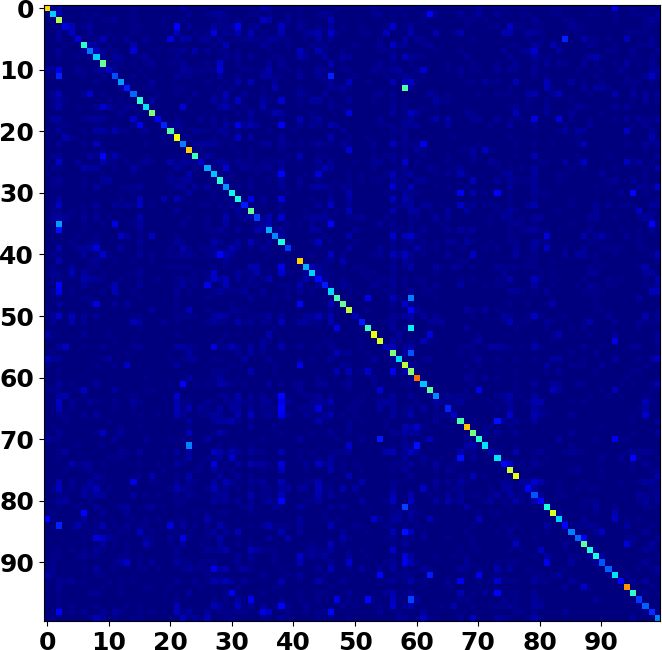}
&\includegraphics[width=.22\textwidth]{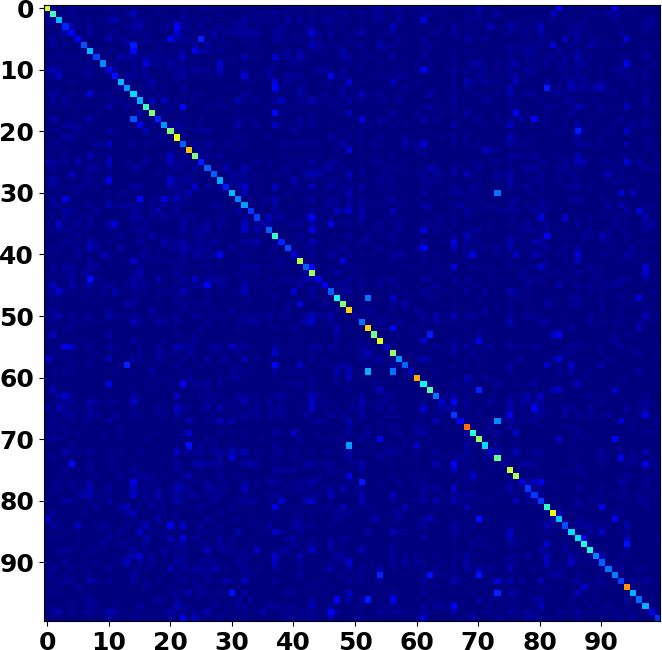}\\
\hline
\multicolumn{2}{c|}{GAC-SNN} &\multicolumn{2}{c}{Legend}\\
\includegraphics[width=.22\textwidth]{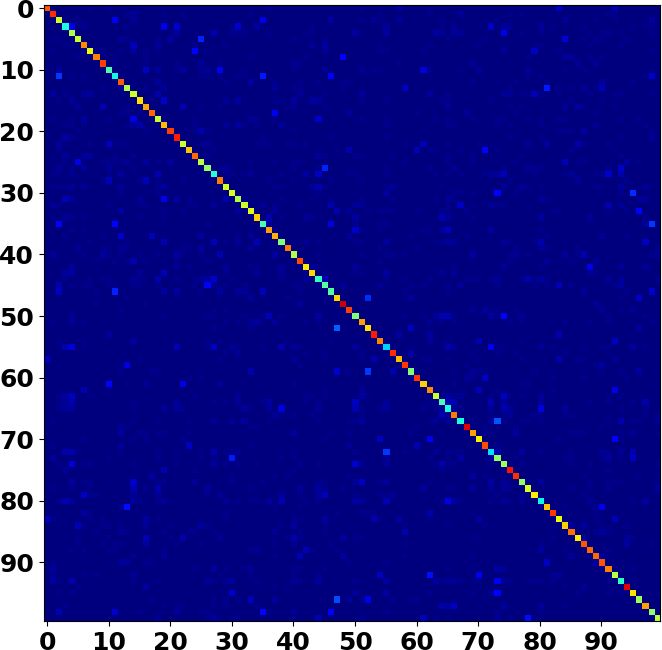}
&\includegraphics[width=.22\textwidth]{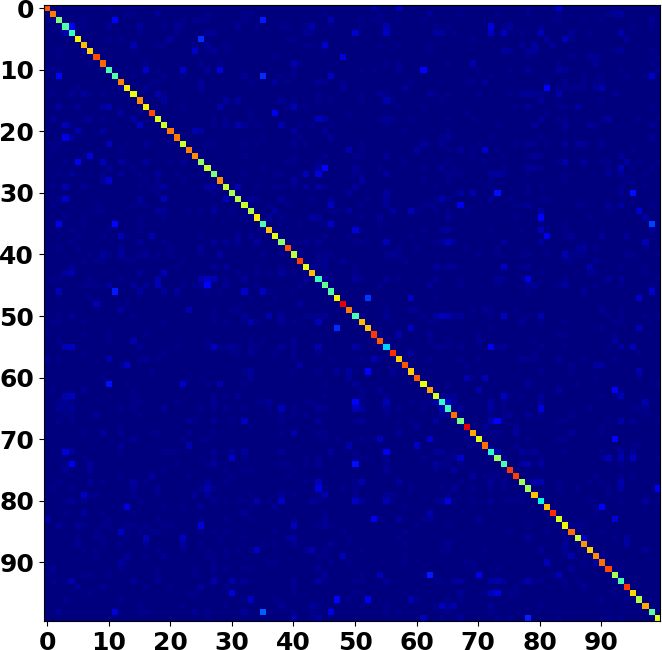}
&\multicolumn{2}{c}{\includegraphics[width=.45\textwidth]{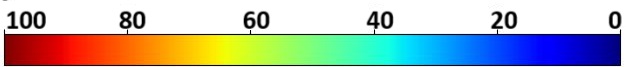}}
\end{tabular}
\caption{Confusion matrix: standard, weighted convolution. ResNet 56, VGG, NiN, gMLP, and GAC-SNN.\label{fig:CLASSRES}}
\end{figure*}
\section{Experimental results}\label{sec:EXPRES}
The training of both the classification and denoising models is performed on the CINECA Leonardo cluster, at the 7th position in the "top500" list~\cite{urlcineca}. The cluster uses 3456 nodes and it is based on BullSequana XH2135 supercomputer nodes, each composed of four accelerators, Nvidia custom Ampere GPU 64GB HBM2 and a single CPU Intel Xeon 8358 with 32 cores (2.60GHz). The module provides 110,592 cores and a Rpeak (i.e., theoretical peak performance) of 304.47 PFlop/s. The parallelisation and high-performance of the training allow us to tune the density function to improve the accuracy of the weighted convolution and the results of the classification (Sect.~\ref{sec:CLASSRES}) and denoising (Sect.~\ref{sec:DENRES}).
\subsection{Classification}\label{sec:CLASSRES}
\paragraph*{Dataset~$\&$ quantitative metrics}
The CIFAR-100 dataset~\cite{krizhevsky2009learning} is a recognised benchmark in computer vision, commonly used for evaluating image classification algorithms. It is composed of 60K colour images at a resolution of~$32\times32$ pixels, evenly distributed across 100 classes, with 600 images per class, and split into 50K training images and 10K test images. The CIFAR-100 covers a wide range of categories, e.g., animals, cars, and fruits. The large number of classes and low resolution of the images make CIFAR-100 a challenging benchmark.

Given \emph{TP} as true positive, \emph{FP} as false positive, \emph{TN} as true negative, and \emph{FN} as false negative prediction, we define the~$\text{accuracy} := (TP + TN) / (TP + TN + FP + FN)$, the~$F1 -\text{score} := TP / (TP + 0.5(FP + FN) )$. Given a classification problem with~$n$ classes, the confusion matrix is the~$n \times n$~$\bm{C}$ matrix where~$\bm{C}_{ij}$ is the number of instances of class~$i$ predicted as class~$j$. The diagonal elements represent the true positives.
\begin{table}[t]
\centering
\captionsetup{width=0.9\textwidth}
\caption{CIFAR-100 classification problem: quantitative metrics. Best results in bold. We report the~$\bm{\alpha}_1$ value.\label{tab:CLASSRES}}
\begin{tabular}{c|c|cc}
\textbf{Method} &~$\bm{\alpha}$ &\textbf{Accuracy} &\textbf{$F1$-score} \\ \hline
\multirow{2}{*}{ResNet 56} & 1.0 &~$39.59\%$ & 0.384 \\
                           & 1.15 &~$\bm{43.97}\%$ &~$\bm{0.434}$ \\ \hline \hline
\multirow{2}{*}{VGG} & 1.0 &$56.89\%$ & 0.566 \\
                                & 0.75 &~$\bm{66.94}\%$ &~$\bm{0.670}$  \\ \hline \hline
\multirow{2}{*}{NiN} & 1.0 &~$51.96\%$ & 0.518 \\
                                & 0.9 &$\bm{52.35}\%$ &~$\bm{0.522}$  \\ \hline \hline
\multirow{2}{*}{gMLP} & 1.0 &~$32.21\%$ & 0.307 \\
                                & 0.95 &$\bm{32.66}\%$ &~$\bm{0.341}$  \\ \hline \hline
\multirow{2}{*}{GAC-SNN} & 1.0 &~$54.32\%$ & 0.540 \\
                                & 0.8 &$\bm{62.24}\%$ &~$\bm{0.624}$  \\ \hline \hline                                                           
\end{tabular}
\end{table}
\paragraph*{Methods and hyper-parameters}
We compare five methods: VGG, ResNet-56, gMLP, NiN, and GAC-SNN. For a fair comparison, we define the same hyper-parametrisation for the five methods: we apply the stochastic gradient descent~\cite{amari1993backpropagation} optimiser with a learning rate of 0.1, a momentum of 0.9, and a decay of~$0.5 \times 10^{-4}$, a cosine annealing of the learning rate, a maximum number of epochs of 200 with an early stopping criterion by monitoring the validation loss, a cross entropy loss function with label smoothing of~$0.1$, a data augmentation with random horizontal flip, a batch size of~$128$. The kernel weights are initialised with the Kaiming initialisation~\cite{he2015delving}, using the same initialisation for both the standard and the weighted convolution. Due to the small resolution of the images, we apply only~$1 \times 1$ and~$3 \times 3$ convolution kernels. In the first case, the density function is not relevant. In the second case, we define the hyperparameter~$\bm{\alpha}_1$ in the range~$(0.5, 1.5)$. Finally, we recall that the training with the \emph{EfficientNet} method did not converge with both standard and weighted convolution. The selected hyperparameterisation might have affected the convergence of the training. 
\begin{table}[t]
\centering
\captionsetup{width=0.9\textwidth}
\caption{CIFAR-100 classification problem: training execution time [seconds] per epoch.\label{tab:CLASSTIME}}
\begin{tabular}{c|cc}
\textbf{Method} &\textbf{Standard convolution} &\textbf{Weighted convolution}\\ \hline
ResNet 56 &202 &215 \\
VGG & 378 &414 \\
NiN & 109 &115 \\
gMLP & 54 &56 \\
GAC-SNN & 118  &130 
\end{tabular}
\end{table}
\paragraph*{Experimental results}
Table~\ref{tab:CLASSRES} shows the accuracy and F1-score of the five methods, comparing standard convolution and weighted convolution. For all the methods, the weighted convolution improves the results in terms of both the quantitative metrics. For example, VGG has an accuracy of~$66.94\%$ with the weighted convolution, compared to~$56.89\%$ with standard convolution. We underline that some methods have a major improvement, due to the larger relevance of the convolution operator inside the learning architecture. Furthermore, the density function can assume values closer to 1 (e.g., in the gMLP method). In this case, the accuracy of standard and weighted convolutions is similar. However, we underline that the density function is always different from the uniform one. This result supports the application of the weighted convolution to improve the accuracy of the classification. Fig.~\ref{fig:CLASSRES} shows the confusion matrix of the five methods, comparing standard convolution (left) with weighted convolution (right) results.

Table~\ref{tab:CLASSTIME} reports the training time per epoch of each model for both standard and weighted convolution. We underline that the weighted convolution has an average of~$ 5\%$ more execution time. In this case, the low resolution of the input images increases the impact of the density function on the convolution operation time.
\begin{figure}[t]
\centering
\begin{tabular}{cc}
\includegraphics[width=.3\textwidth]{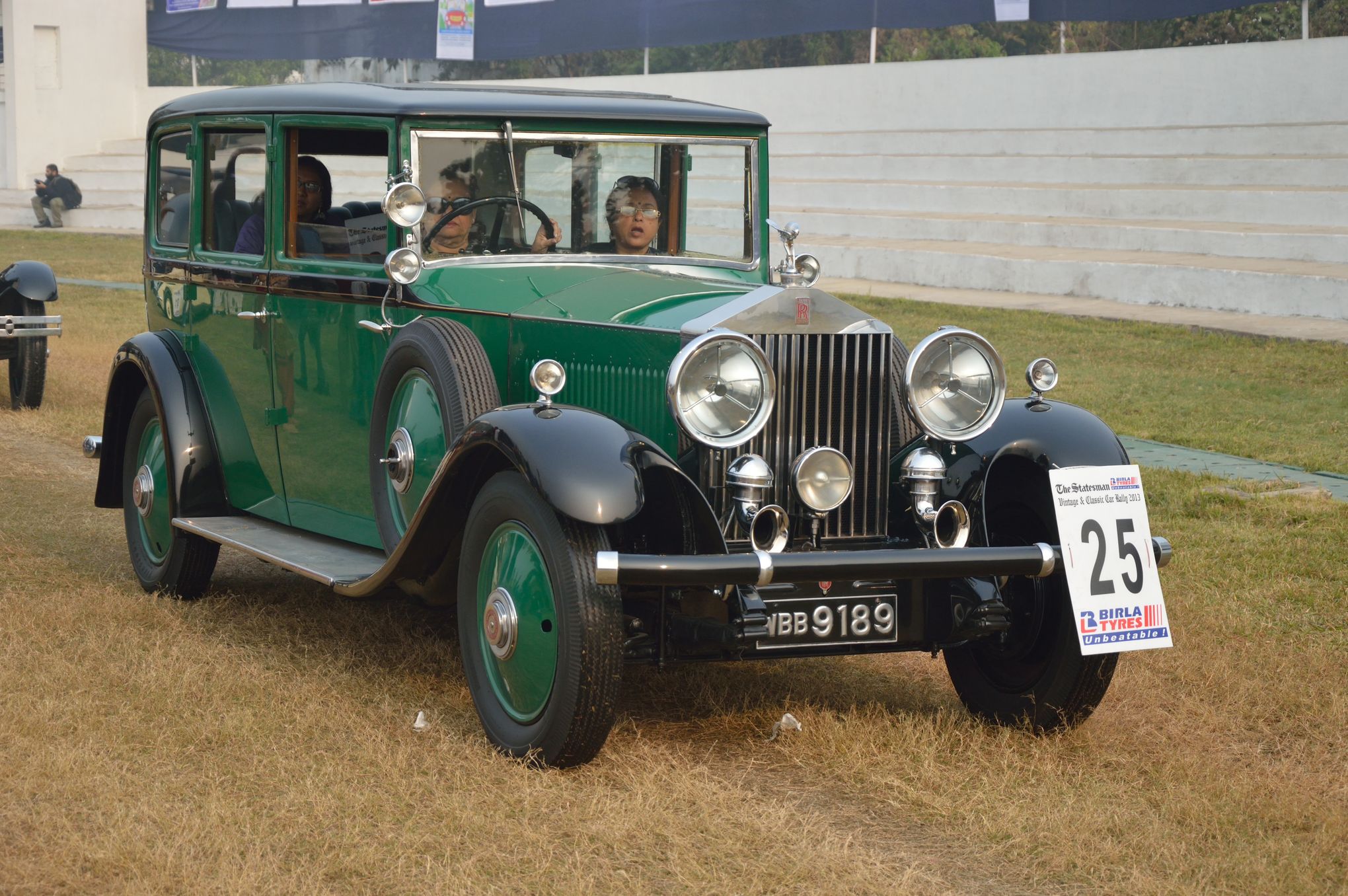} & \hspace{-4mm}
\includegraphics[width=.3\textwidth]{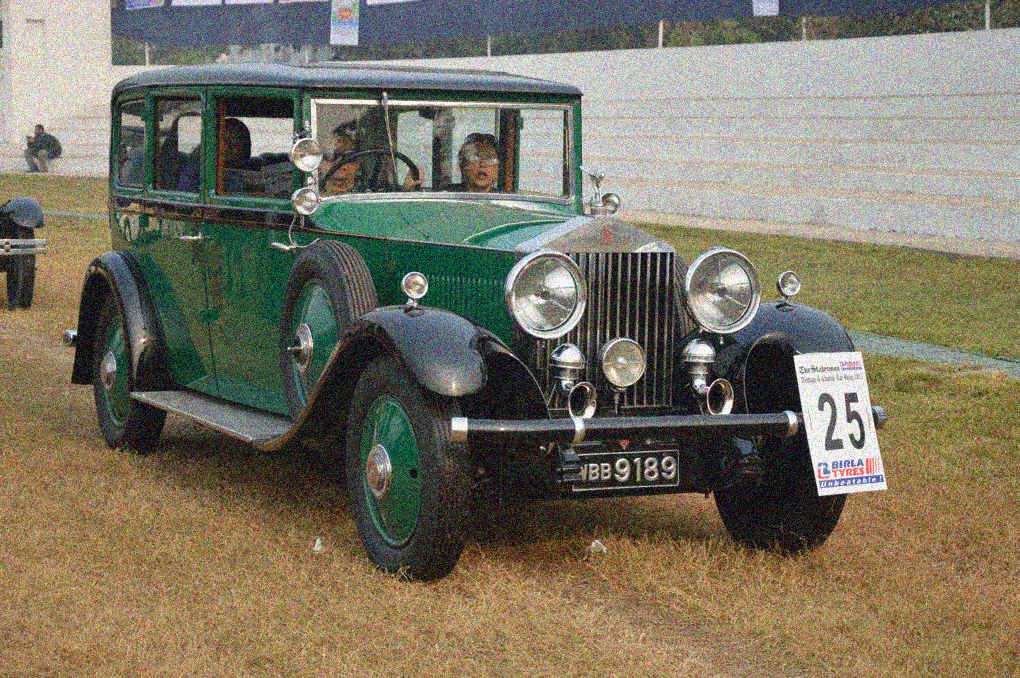} \\
Ground-truth & Noisy  \\ \hline \hline
\end{tabular}
\vspace{2mm}
\begin{tabular}{ccc}
\includegraphics[width=.3\textwidth]{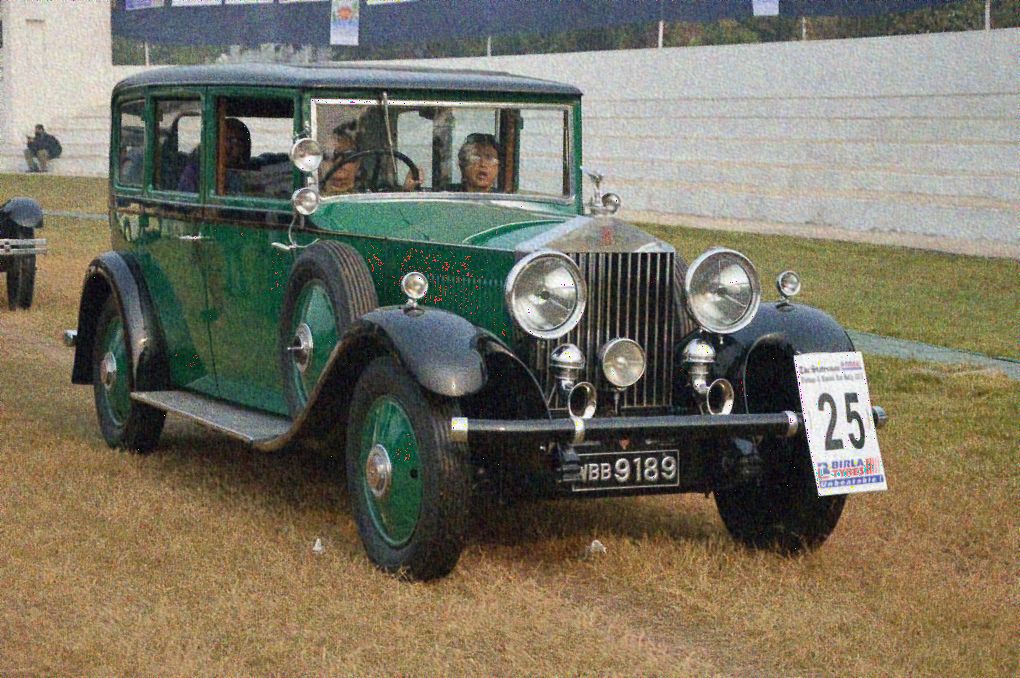} & \hspace{-4mm}
\includegraphics[width=.3\textwidth]{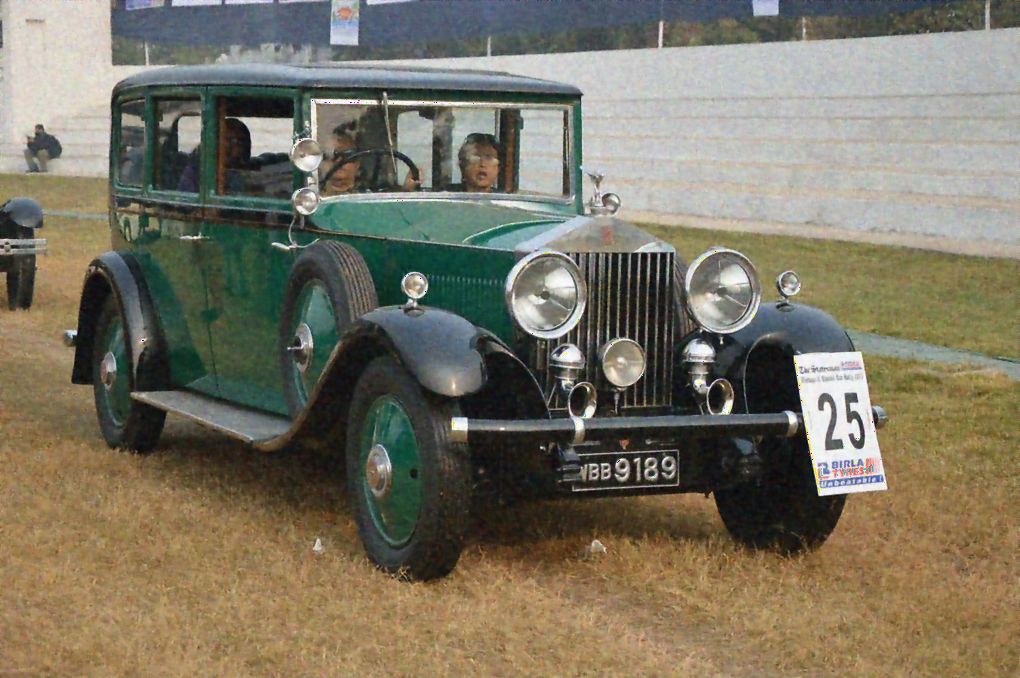} & \hspace{-4mm}
\includegraphics[width=.3\textwidth]{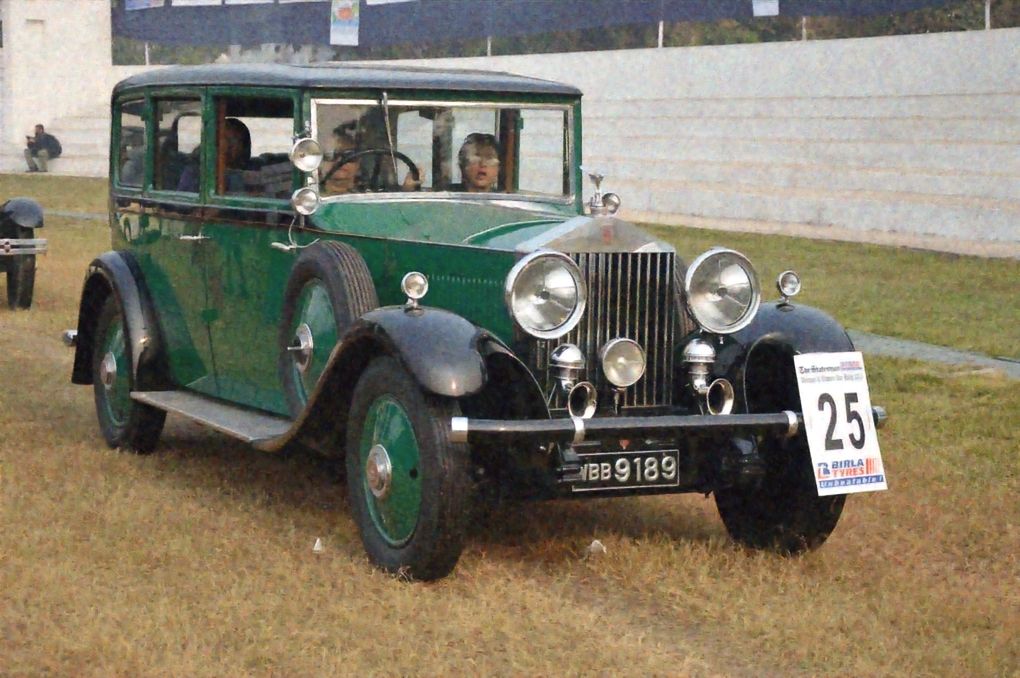} \\
\includegraphics[width=.3\textwidth]{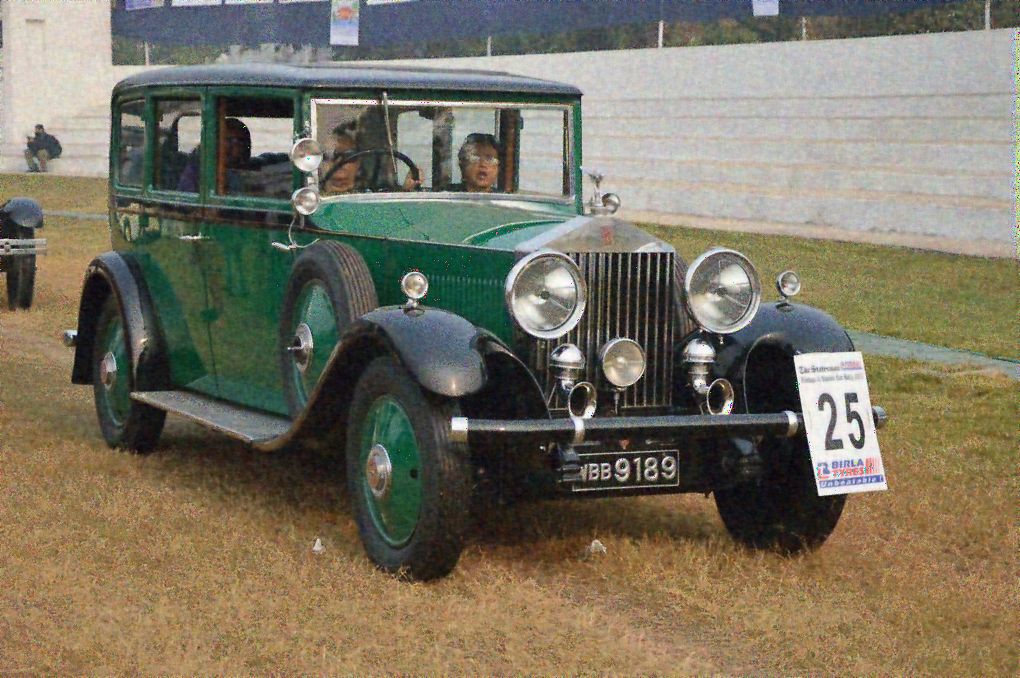} & \hspace{-4mm}
\includegraphics[width=.3\textwidth]{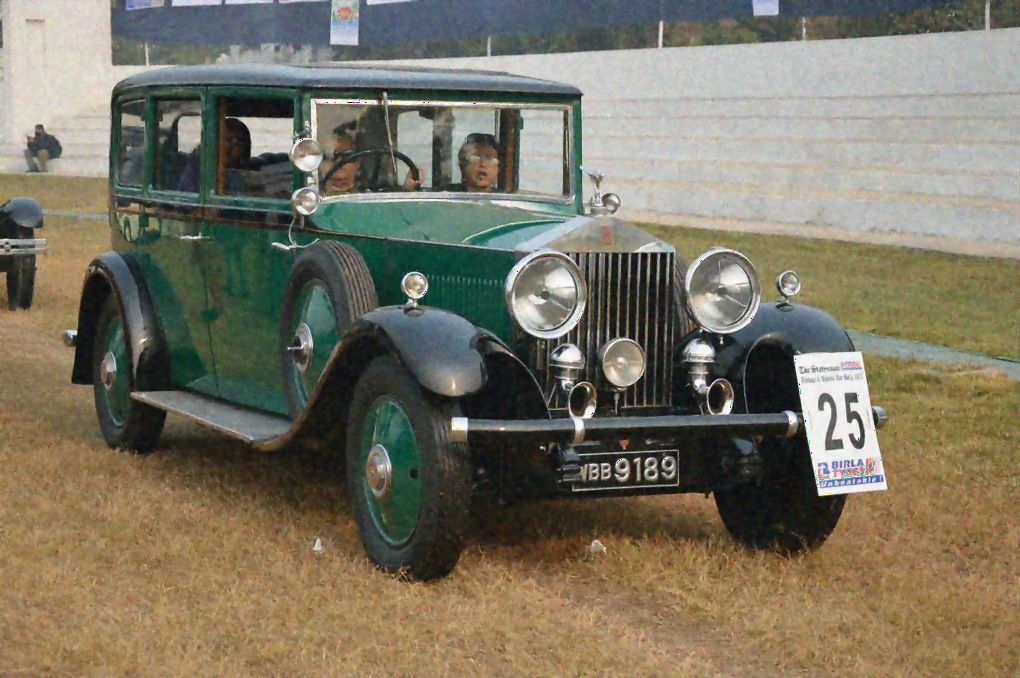} & \hspace{-4mm}
\includegraphics[width=.3\textwidth]{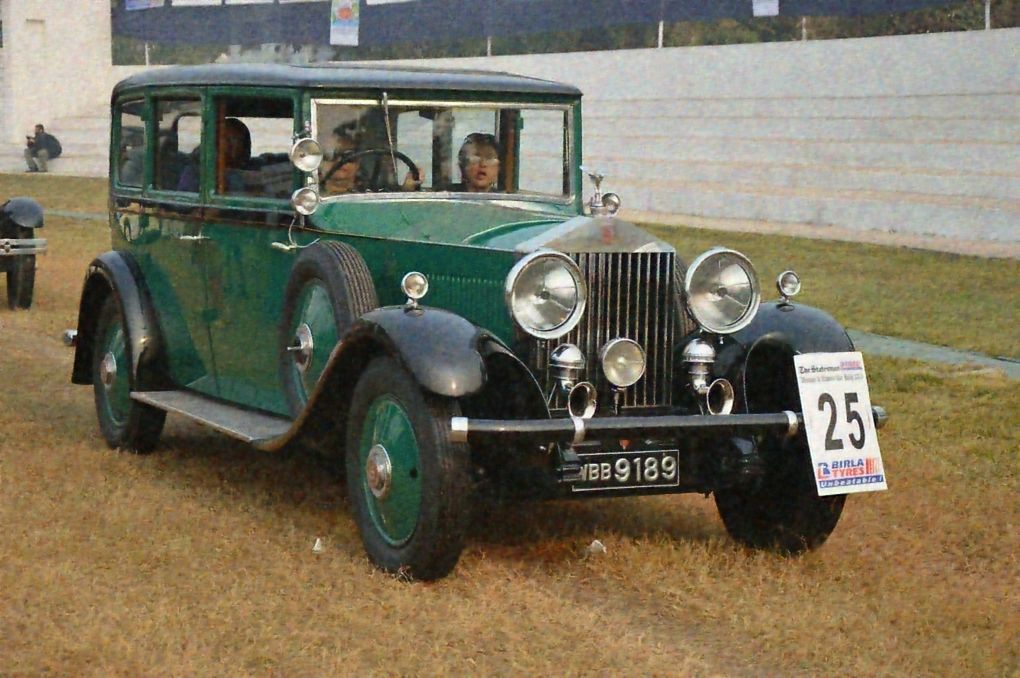} \\
DnCNN & NAFNet & CGNet
\end{tabular}
\caption{Denoising results with standard (second row) and weighted (third row) convolution with~$3 \times 3$ kernel.}
\label{fig:DENRES1}
\end{figure}
\subsection{Denoising}\label{sec:DENRES}
\paragraph*{Dataset~$\&$ quantitative metrics}
The DIV2K dataset~\cite{Agustsson2017CVPRWorkshops} is a benchmark widely used for image super-resolution and restoration tasks. It is composed of 2K RGB images with resolutions close to~$2,048 \times 1,080$ and a variety of natural scenes and objects. The dataset is split into 800 images for training, 100 for validation, and 1,100 for testing. DIV2K is a standard benchmark for image denoising and super resolution algorithms, allowing the user to evaluate the quality of textures, geometries, and artefacts removal.
\begin{figure}[t]
\centering
\begin{tabular}{cc}
\includegraphics[width=.3\textwidth]{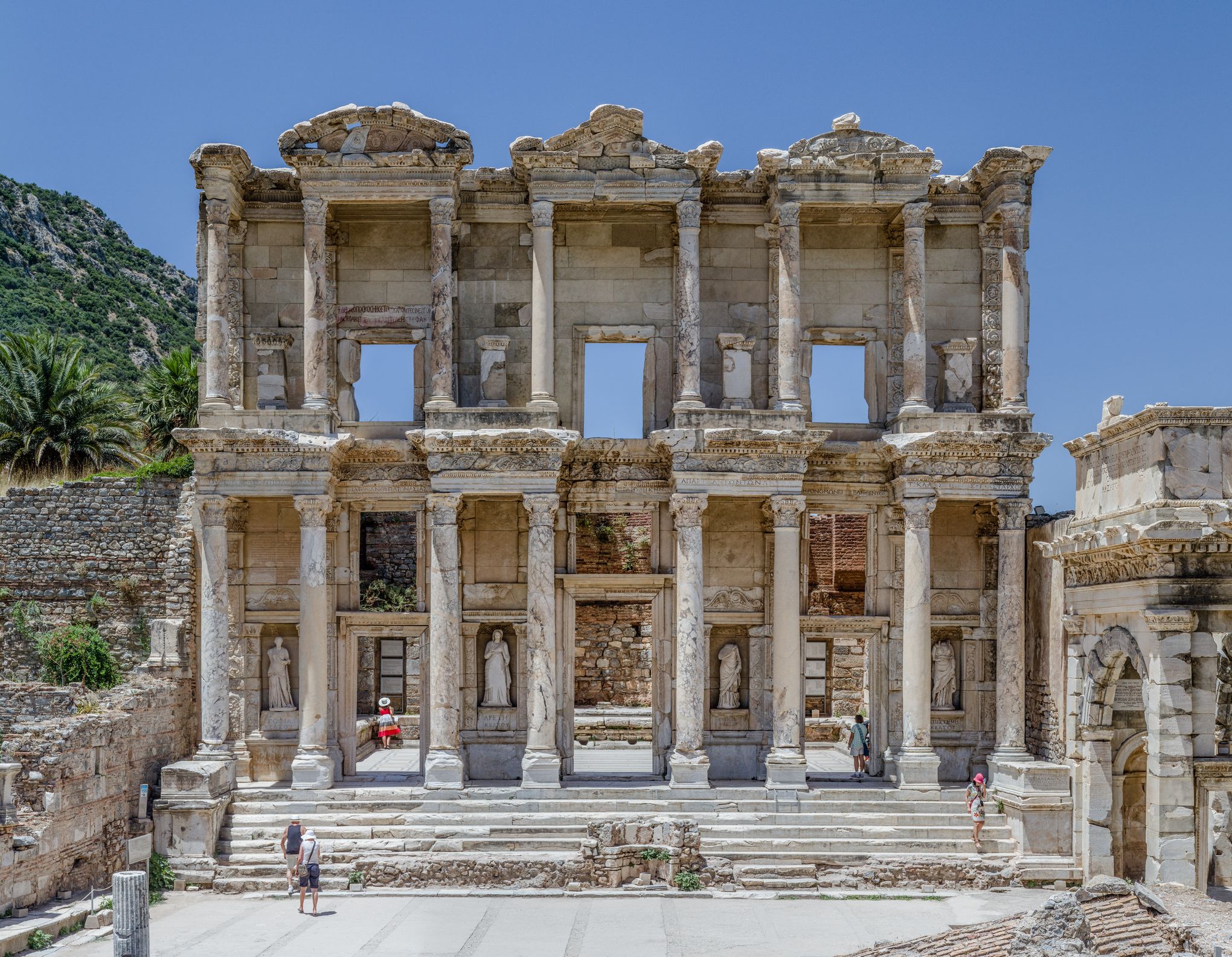} & \hspace{-4mm}
\includegraphics[width=.3\textwidth]{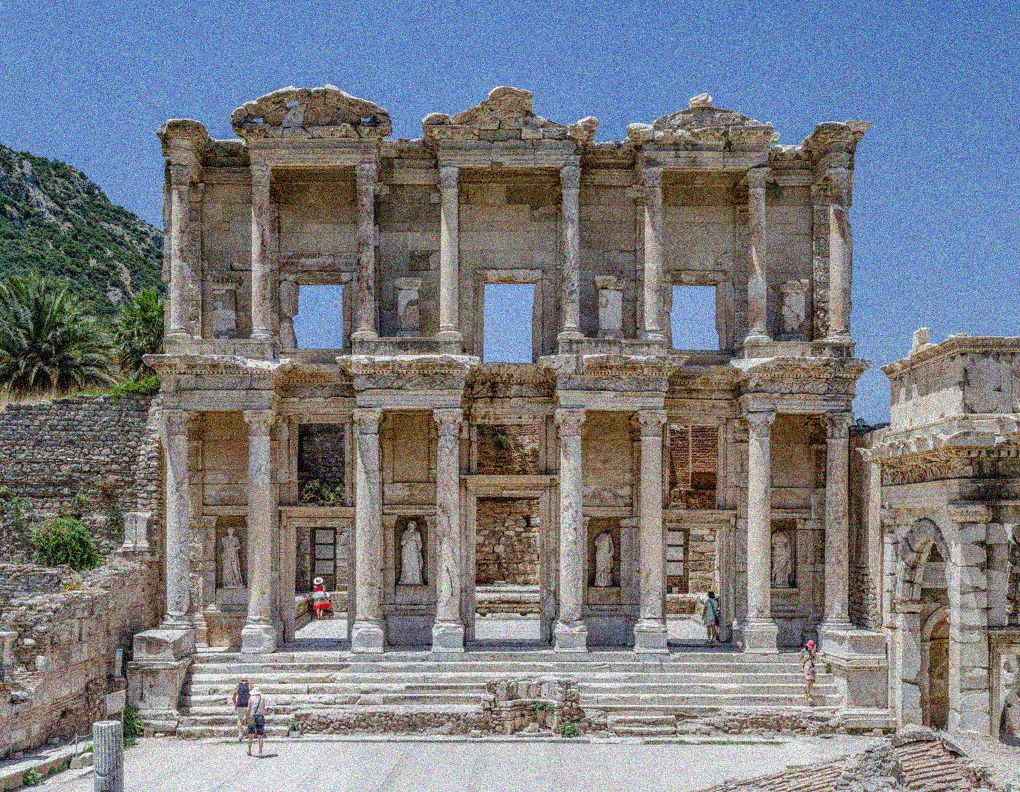} \\ \hline \hline
\end{tabular}
\vspace{2mm}
\begin{tabular}{ccc}
\includegraphics[width=.3\textwidth]{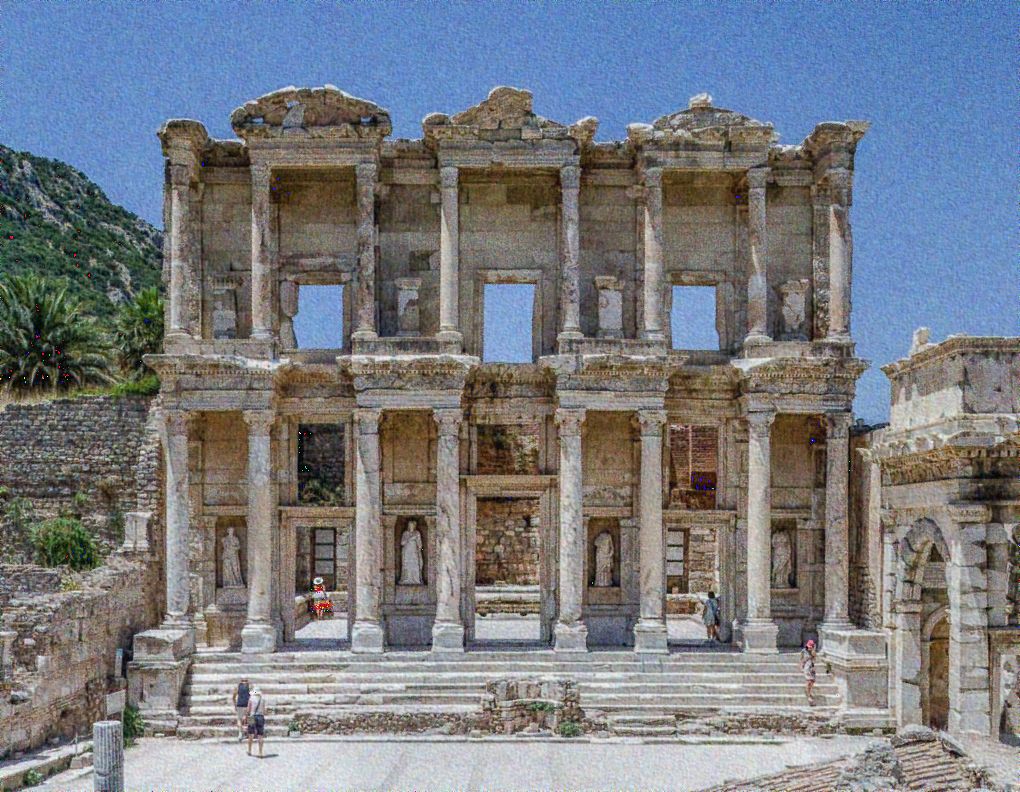} & \hspace{-4mm}
\includegraphics[width=.3\textwidth]{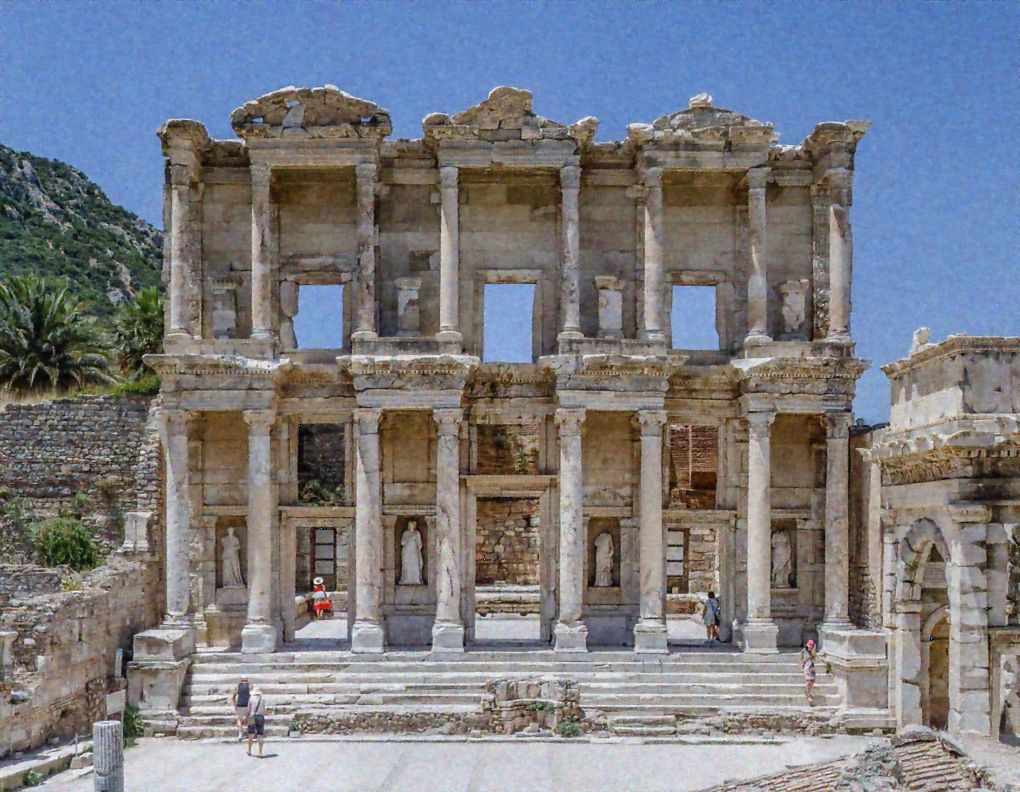} & \hspace{-4mm}
\includegraphics[width=.3\textwidth]{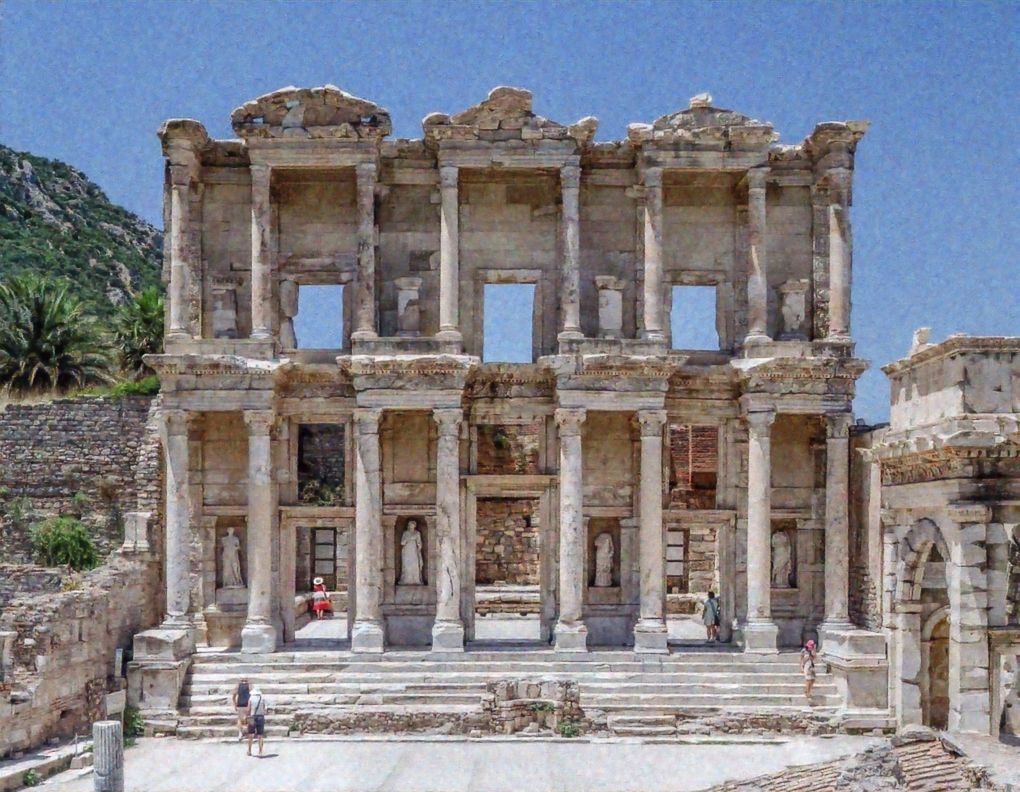} \\
\includegraphics[width=.3\textwidth]{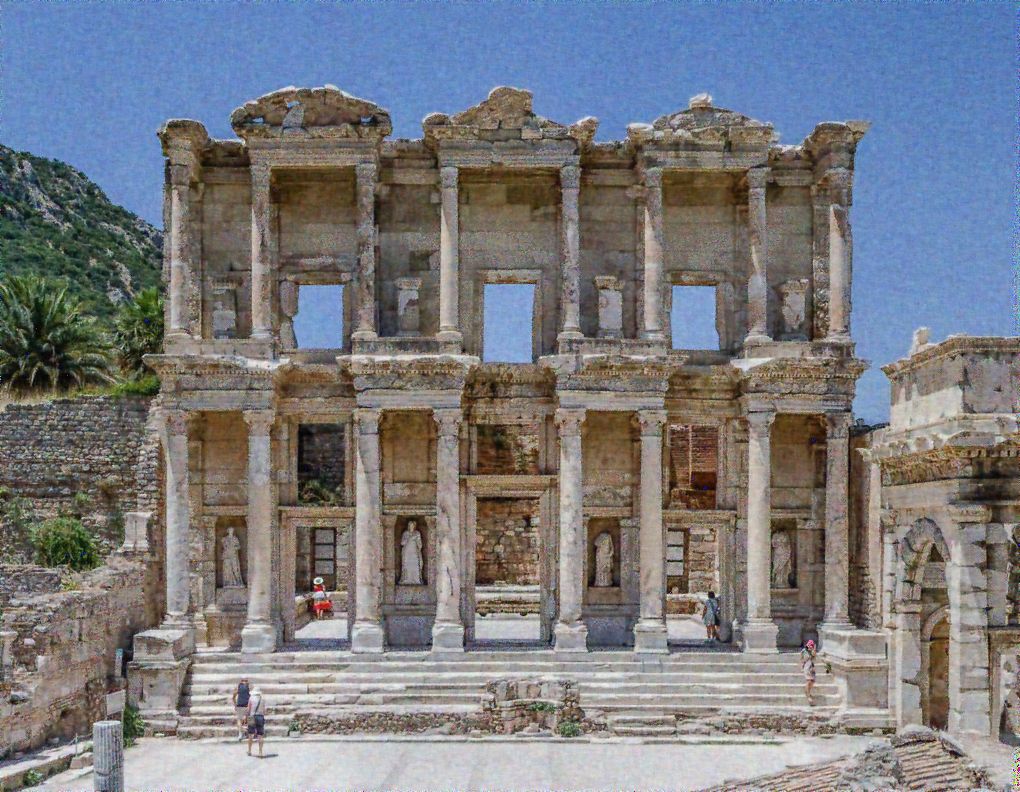} & \hspace{-4mm}
\includegraphics[width=.3\textwidth]{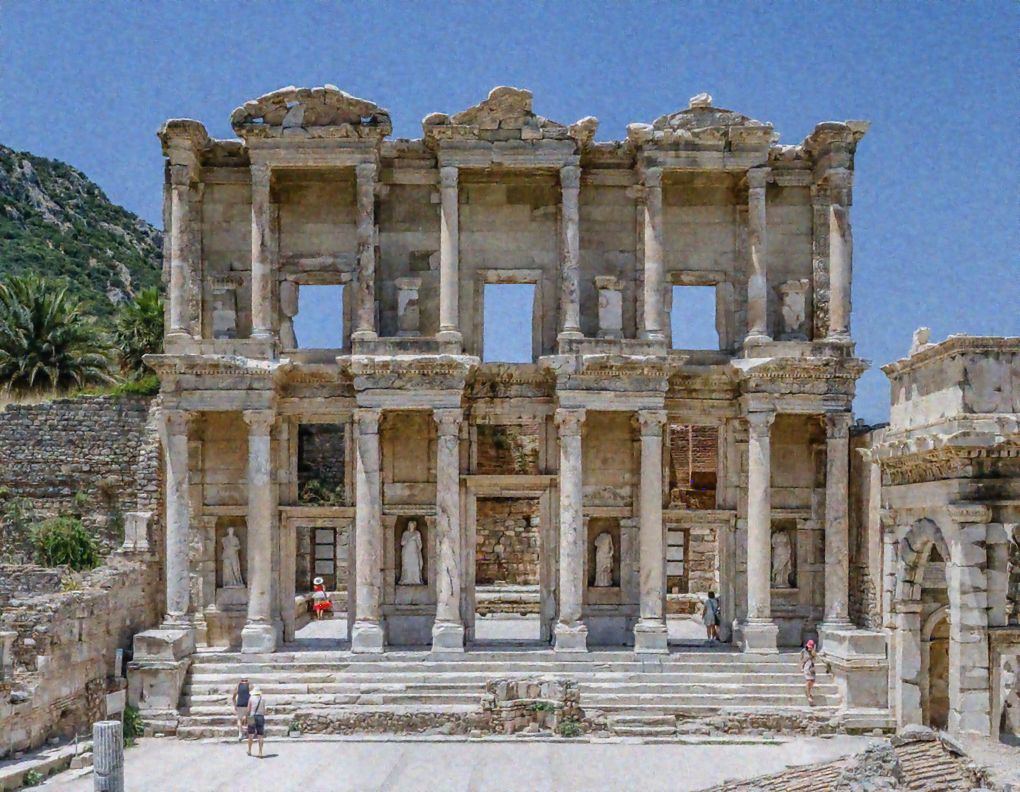} & \hspace{-4mm}
\includegraphics[width=.3\textwidth]{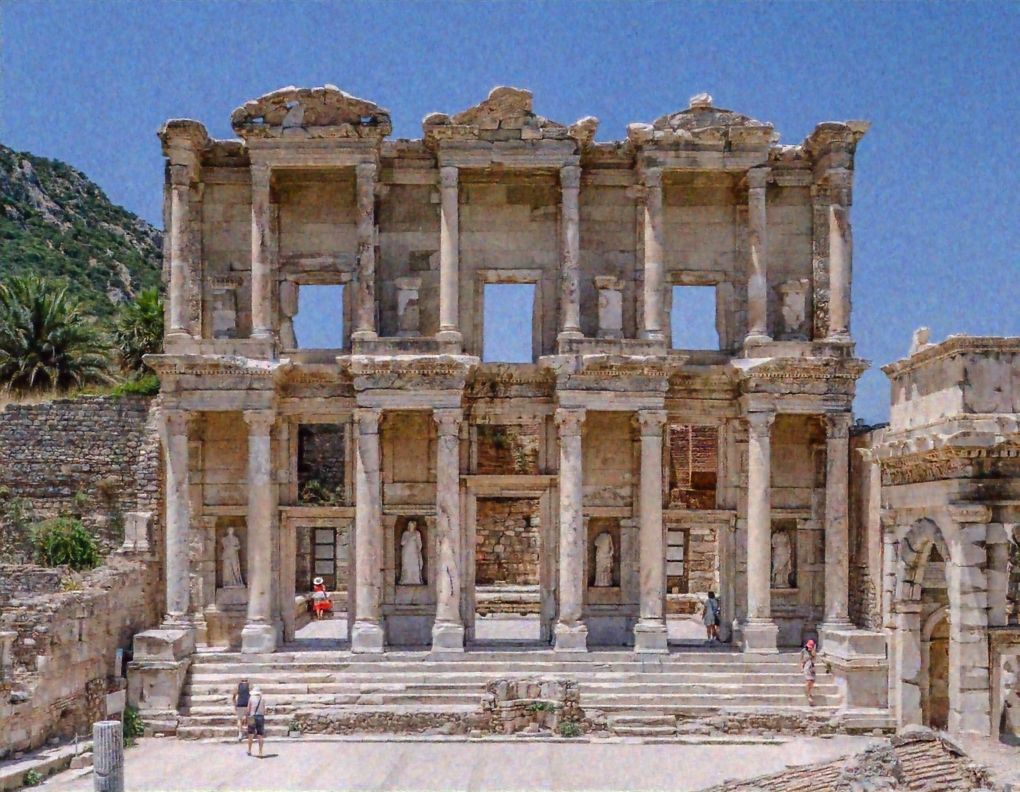} \\
DnCNN & NAFNet & CGNet
\end{tabular}
\caption{Denoising results with standard (second row) and weighted (third row) convolution with~$5 \times 5$ kernel.}
\label{fig:DENRES2}
\end{figure}

We quantify the results of the denoising by comparing the predicted and the ground-truth images through several metrics for local, texture, noise-based, and structural similarity: \emph{normalised mean squared error} (NRMSE), \emph{peak-signal-to-noise-ratio} (PSNR), \emph{structural similarity} (SSIM), \emph{universal image quality}~\cite{wang2002universal} (UIQ), \emph{feature-based similarity}~\cite{zhang2011fsim} (FSIM). SSIM and FSIM vary from 0 (worst case) to 1 (best case), PSNR varies from 0 (worst case) to~$+\infty$ (best case), NRMSE varies from~$+\infty$ (worst case) to 0 (best case), UIQ varies from~$-1$ (worst case) to~$+1$ (best case).
\begin{table}[t]
\centering
\captionsetup{width=0.9\textwidth}
\caption{DIV2K denoising: quantitative metrics. Best results in bold. We report the~$\bm{\alpha}_1$ value for the~$3 \times 3$ kernel and the~$\bm{\alpha}_1, \bm{\alpha}_2$ values for the~$5 \times 5$ kernel. \label{tab:DENRES}}
\begin{tabular}{c|c|c|ccccc}
\textbf{Kernel} &\textbf{Method}  &~$\bm{\alpha}$ &\textbf{NRMSE} &\textbf{PSNR} &\textbf{SSIM} &\textbf{FSIM} &\textbf{UIQ} \\ \hline
\multirow{6}{*}{$3 \times 3$}
& \multirow{2}{*}{DnCNN} &  1.0 & 0.112 & 20.17 & 0.735 & 0.921 & 0.925 \\
& &  0.8 &$\bm{0.081}$ &~$\bm{22.63}$ &~$\bm{0.826}$ &~$\bm{0.942}$ &~$\bm{0.937}$ \\
\hhline{~|-------}
& \multirow{2}{*}{NAFNet} &  1.0 & 0.064 & 24.82 & 0.865 & 0.963 & 0.938 \\
& & 0.7 &$\bm{0.055}$ &~$\bm{25.83}$ &~$\bm{0.881}$ &~$\bm{0.969}$ &~$\bm{0.940}$ \\
\hhline{~|-------}
& \multirow{2}{*}{CGNet} & 1.0 &0.048 & 26.57 & 0.896 & 0.970 & 0.933 \\
& &  0.8 &$\bm{0.045}$ &~$\bm{28.01}$ &~$\bm{0.901}$ &~$\bm{0.969}$ &~$\bm{0.932}$ \\ \hline \hline

\multirow{6}{*}{$5 \times 5$}
& \multirow{2}{*}{DnCNN} &  (1.0, 1.0) &0.266 & 12.15 & 0.402 & 0.768 & 0.813 \\
& &  (0.1, 0.9) &$\bm{0.074}$ &~$\bm{23.35}$ &~$\bm{0.830}$ &~$\bm{0.943}$ &~$\bm{0.938}$ \\
\hhline{~|-------}
& \multirow{2}{*}{NAFNet} &  (1.0, 1.0)  &0.068 & 24.13 & 0.842 & 0.957 & 0.934 \\
& &  (0.5, 0.9) &$\bm{0.047}$ &~$\bm{27.19}$ &~$\bm{0.918}$ &~$\bm{0.974}$ &~$\bm{0.941}$ \\
\hhline{~|-------}
& \multirow{2}{*}{CGNet} &  (1.0, 1.0)  & 0.051 & 26.01 & 0.801 & 0.969 & 0.932 \\
& &  (0.6, 0.9) &$\bm{0.044}$ &~$\bm{28.07}$ &~$\bm{0.902}$ &~$\bm{0.969}$ &~$\bm{0.934}$
\end{tabular}
\end{table}
\paragraph*{Experimental results}
We compare three methods on the test dataset: DnCNN, NAFNet, and CGNet. For a fair comparison, we define the same hyper-parametrisation for the three methods: we apply the Adam~\cite{kingma2014adam} optimiser with~$\beta_1 = 0.9, \beta_2 = 0.999$, mean squared error loss function, 100 epochs with an early stopping criterion by monitoring the validation loss, learning rate of 0.1 with cosine annealing. The kernel weights are initialised with the Kaiming initialisation~\cite{he2015delving}, using the same initialisation for both the standard and the weighted convolution. We apply a Gaussian noise with~$\mu = 0, \sigma = 0.01$, and we use the same input and noisy images for all the methods. We apply both~$3 \times 3$ and~$5 \times 5$ convolution kernels. In the first case, we define the hyperparameter~$\bm{\alpha}_1$ in the range~$(0.5, 1.5)$. In the second case, we define the hyper-parameter~$\bm{\alpha}_1$ in the range~$(0.05, 1)$ and ~$\bm{\alpha}_2$ in the range~$(0.5, 1.5)$.

Table~\ref{tab:DENRES} shows the quantitative metrics for the three methods with both standard and weighted convolution and with~$3 \times 3$ and~$5 \times 5$ kernels. The weighted convolution improves the standard convolution under every metric and for all the methods. For example, DnCNN has an SSIM value with the weighted convolution of~$0.826$, versus an SSIM value of~$0.735$ with the standard convolution, with a~$3 \times 3$ kernel. CGNet has a PSNR value with the weighted convolution of~$28.07$, versus a PSNR value of~$26.01$ with the standard convolution, with a~$5 \times 5$ kernel. We also underline that the weighted convolution with~$5 \times 5$ kernel size and weighted convolution has better results than both standard and weighted convolution with~$3 \times 3$ kernel size. For example, NAFNet with~$5 \times 5$ kernel size and~$\bm{\alpha} = (0.5, 0.9, 1.0, 0.9, 0.5)$ has a PSNR value of 27.19, compared to the~$3 \times 3$ kernel where the PSNR value is 25.83 and 24.82 with the weighted ($\bm{\alpha} = (0.7, 1.0, 0.7)$) and standard ($\bm{\alpha} = (1.0, 1.0, 1.0)$) convolution, respectively.

Figs.~\ref{fig:DENRES1},~\ref{fig:DENRES2} show the denosing results for the three methods, with standard and weighted convolution,~$3 \times 3$ and~$5 \times 5$ kernel size. In Fig.~\ref{fig:DENRES1}, the DnCNN method, weighted convolution reduces the artefacts (e.g. the red dots on the car) with respect to the standard convolution. In Fig.~\ref{fig:DENRES2}, the weighted convolution slightly improves the smoothing level, e.g., in the background of the image. Table~\ref{tab:DENTIME} reports the training time per epoch of each model with both standard and weighted convolution. In this case, the large resolution of the input images reduces the impact of the density function on the convolution operation time.

\section{Conclusions and future work}\label{sec:CONC}
We have developed an efficient implementation of the weighted convolution through the computation of kernel weights with a density function. We have tested our implementation with classification and denoising problems, comparing state-of-the-art deep learning methods on standard benchmarks. Our weighted convolution has better results than standard convolution for all the methods under different quantitative metrics. We have performed the experimental tests on the CINECA Leonardo cluster, whose high performance allowed us to tune the values of the density function to improve the accuracy of the denoising and classification. As future work, we plan to extend the weighted convolution to 3D kernels and apply the weighted convolution to applicative contexts, e.g., 2D and 3D biomedical data, for denoising, super-resolution, and classification tasks.
\begin{table}[t]
\centering
\captionsetup{width=0.9\textwidth}
\caption{DIV2K denoising problem: training execution time [seconds] per epoch.\label{tab:DENTIME}}
\begin{tabular}{c|c|cc}
\textbf{Kernel size} &\textbf{Method} &\textbf{Standard convolution} &\textbf{Weighted convolution} \\ \hline
\multirow{3}{*}{$3\times3$}&DnCNN &2.43 &2.50 \\
&NAFNET &2.08 &2.10 \\
&CGNet & 3.04 & 3.07  \\ \hline \hline
\multirow{3}{*}{$5\times5$} &DnCNN &9.48 &9.76 \\
&NAFNET &8.92 &9.11 \\
&CGNet & 11.02 & 11.13
\end{tabular}
\end{table}

\textbf{Acknowledgements}
SC and GP are part of RAISE Innovation Ecosystem, funded by the European Union - NextGenerationEU and by the Ministry of University and Research (MUR), National Recovery and Resilience Plan (NRRP), Mission 4, Component 2, Investment 1.5, project “RAISE - Robotics and AI for Socio-economic Empowerment” (ECS00000035) and the PRIN 2022 Project 2022WK7NHC. Tests on the CINECA Cluster are supported by the ISCRA-C project US-SAMP, HP10CXLQ1S.

\bibliographystyle{alpha}
\bibliography{bibliography}

\end{document}